\pdfoutput=1

\documentclass[11pt]{article}

\usepackage{latex/acl}

\usepackage{times}
\usepackage{latexsym}

\usepackage[T1]{fontenc}

\usepackage[utf8]{inputenc}

\usepackage{microtype}
\usepackage{inconsolata}
\usepackage{booktabs}
\usepackage{graphicx}
\usepackage{subcaption}
\usepackage{float}
\graphicspath{ {./files/} }

\def\ANDNOBF{\end{tabular}\hss\egroup \hfil\hfil\egroup
          \vskip 0.25in plus 1fil minus 0.125in
           \hbox to \linewidth\bgroup\large \hfil\hfil
             \hbox to 0pt\bgroup\hss \begin{tabular}[t]{c}}

\title{NextLevelBERT: Masked Language Modeling with Higher-Level Representations for Long Documents} 

\author{Tamara Czinczoll \And Christoph Hönes \And Maximilian Schall \And Gerard de Melo \ANDNOBF Hasso Plattner Institute / University of Potsdam\\ Potsdam, Germany \\{\texttt{\url{https://github.com/aiintelligentsystems/next-level-bert}}}\\ \texttt{tamara.czinczoll@hpi.de}}

\begin{document}
\maketitle
\begin{abstract}
While (large) language models have significantly improved over the last years, they still struggle to sensibly process long sequences found, e.g., in books, due to the quadratic scaling of the underlying attention mechanism. To address this, we propose NextLevelBERT, a Masked Language Model operating not on tokens, but on higher-level semantic representations in the form of text embeddings. We pretrain NextLevelBERT to predict the vector representation of entire masked text chunks and evaluate the effectiveness of the resulting document vectors on three types of tasks: 1)~Semantic Textual Similarity via zero-shot document embeddings, 2)~Long document classification, 3)~Multiple-choice question answering. We find that next-level Masked Language Modeling is an effective technique to tackle long-document use cases and can outperform much larger embedding models as long as the required level of detail of semantic information is not too fine. Our models and code are publicly available online.
\end{abstract}

\section{Introduction}
\label{sec:intro}

\begin{figure*}[t]
    \centering
    \resizebox{1\linewidth}{!}{
    \includegraphics{{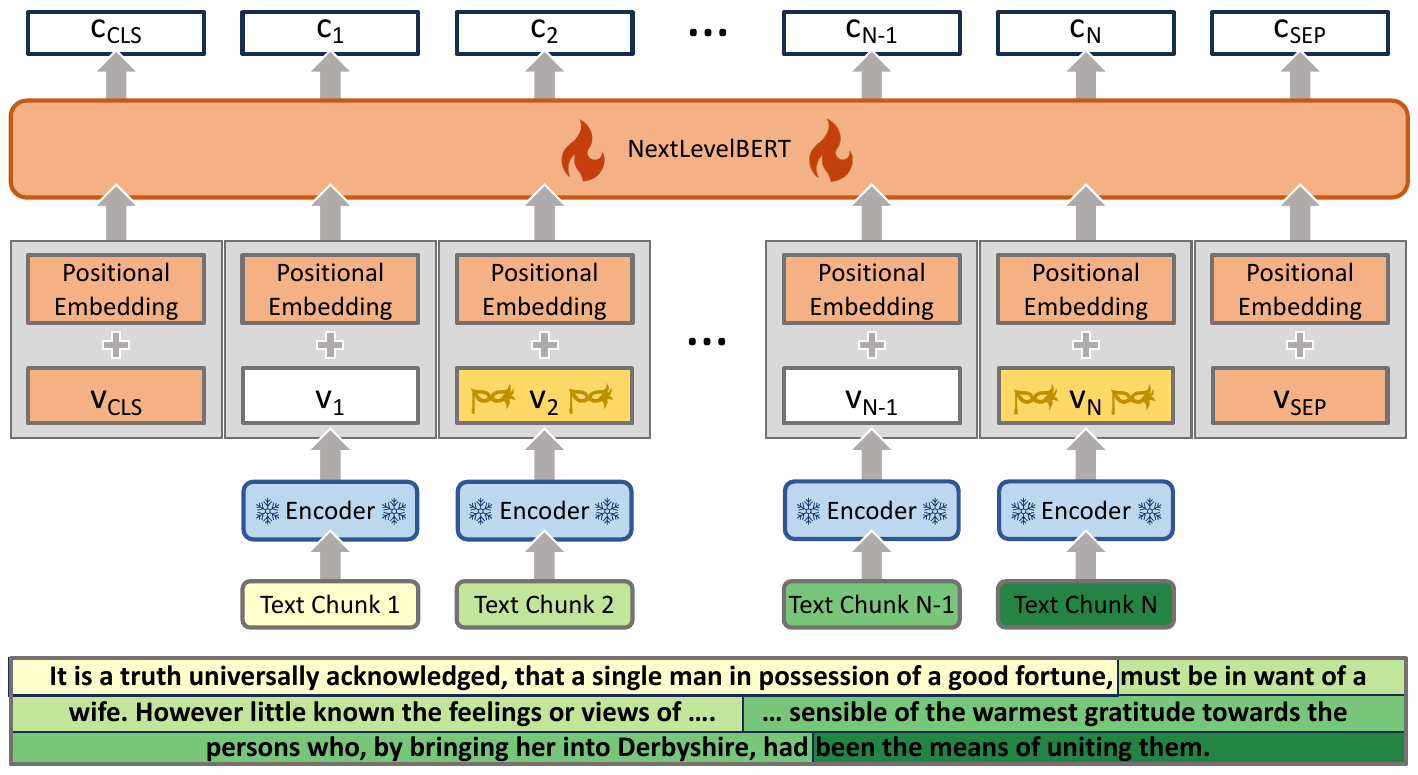}}
    }
    \caption{NextLevelBERT's architecture. In a hierarchical setup, the frozen sentence encoder first generates independent chunk vectors from the input document. The chunk vectors $v_2$ and $v_n$ are masked and need to be predicted from the surrounding context. Through its higher-level Masked Language pretraining objective, NextLevelBERT learns to produce contextualized versions of the chunk vectors.}
    \label{fig:architecture}
\end{figure*}

While the Transformer model is the basis for most large language models,
its attention mechanism scales quadratically with respect to the length of the input sequence 
\citep{vaswani_attention_2017}. This makes it prohibitive to process longer input sequences such as books or movie scripts. 
To address this limitation, the most common strategy has been to sparsify or approximate the attention operation \citep{beltagy_longformer_2020,zhang_poolingformer_nodate,kitaev_reformer_2019,han_hyperattention_2023}, or to extend the positional information during inference \citep{press_train_2022, chen_extending_2023}.
Other approaches to process long documents include switching to a retrieval augmentation approach \cite{RAG}, modeling long-range dependencies via a graph structure \citep{zhang_hegel_2022, bugueno-de-melo-2023-connecting}, as well as compressing the input either via an explicit memory module \citep{rae_compressive_2019, zemlyanskiy_readtwice_2021} or via hierarchical processing \citep{pappagari_hierarchical_2019, ivgi_efficient_2022}. Further related work is discussed in \autoref{sec:related_work}.

From a cognitive perspective, language is widely regarded as a hierarchical process \citep{lashley_problem_1951, davis_hierarchical_2003, everaert_structures_2015}, enabling connections between distant pieces of information. When investigating the differences between how human brains and 
typical large language models 
process language, \citet{caucheteux_evidence_2023} found that humans seem to use a predictive coding hierarchy \citep{friston_predictive_2009} to anticipate words at multiple time spans. 

Beyond word-level predictions, the original BERT architecture \cite{devlin-etal-2019-bert} also includes next sentence prediction as one of the pretraining tasks, but this still is based on individual tokens as input. Masked language modeling has previously been explored at the level of short spans of tokens \cite{joshi-etal-2020-spanbert}, but again assuming tokens as inputs.

We present NextLevelBERT, a masked language model operating entirely at larger timescales, with inputs that are not individual tokens but rather generic chunks of text (e.g., sentences or larger spans of tokens). For each chunk, we obtain a representation using a pretrained encoding model. Some of these chunks are then masked and the respective representations need to be predicted based on the context.
Our results show that applying masked language modeling to higher-level semantic representations results in meaningful document vectors. NextLevelBERT uses only a third of the number of parameters of its strongest contender, but showcases stronger performance than other commonly used approaches during downstream testing as long as the task does not require a high level of semantic detail at the token level.

Our contributions are as  follows: We apply masked language modeling in a novel way on text embeddings and pretrain NextLevelBERT on a sizable collection of very long documents (average document length: $\sim$125,000 tokens). We obtain a hierarchical model that can 1) generate contextualized chunk vectors and 2) aggregate these into meaningful long-document representations. We demonstrate NextLevelBERT's effectiveness on three diverse tasks from the long-document domain and investigate the impact of masking strategy and model initialization during pretraining.

\section{Next-Level Masked Language Modeling}
Our goal is to obtain a masked language model that operates on higher-level semantic units.
Instead of predicting masked tokens based on other tokens in the context, we predict masked text chunk representations based on other text chunk representations. 

\subsection{Design Considerations}

As a starting point, we consider a basic RoBERTa model \citep{liu_roberta_2019}, which we then adapt to operate on generic text vectors instead of tokens.
Analogously to the original RoBERTa, we disregard BERT's original Next Sentence Prediction task and only train the model with a masked language modeling (MLM) objective.

The original MLM objective in BERT and RoBERTa assumes that some randomly chosen tokens are masked or replaced by random tokens.
The model is trained to predict the original tokens at these positions. To succeed, the model needs to take the surrounding tokens as context into account as well as learn general syntactic and semantic patterns. We hypothesize that pretraining with an MLM objective but generalized to generic vector representations will learn similar patterns on a semantically more abstract level.

The NextLevelBERT architecture is illustrated in \autoref{fig:architecture} and explained in detail in the next section.

\subsection{NextLevelBERT Architecture}

\paragraph{Input Encoding} Given a document we first split it into smaller chunks of text, either at the level of sentences or as fixed-length chunks, typically of size 16 to 512 tokens.
Each text chunk is encoded independently with a pretrained encoder-only language model, so that we obtain a sequence of vectors. We encode all documents in this manner, keeping the pretrained encoder's weights frozen.

For each input sequence, we insert a trainable [CLS] vector in the first position and then pack multiple documents separated by a [SEP] vector together until the model's maximum sequence length is reached.

\paragraph{Masking} We then perform masking by randomly picking
each text chunk vector with a probability of $15\%$.
Out of the selected vectors, we replace 80\% with a [MASK] sentence vector, 10\% with a random sentence vector from another sample in the minibatch, and leave the last 10\% unchanged. This is identical to RoBERTa's masking procedure, except that our representations are not regular token-level ones.
The embeddings for the special [MASK], [CLS], and [SEP] tokens are randomly initialized before training the model. 

\paragraph{Prediction} We obtain predictions for the masked sentence objective from an MLP-based task head. In this masked language modeling setup, the model's objective is to predict the original input vector. While regular vocabularies have a fixed size, our model needs to cope with a potentially infinite inventory of sentence embeddings. 
Our model thus is trained to predict the entire original $d$-dimensional chunk vector. Accordingly, instead of the usual cross-entropy loss, we incorporate a Smooth L1 loss,
which worked best in initial experiments.

\paragraph{Discussion} With the typical input length of pretrained Transformer-based language models of 512, as well as our own model's input length of 512, the overall receptive field of the model can be stretched to a maximum of $512 \times 512 =$ 262,144 tokens. Theoretically, this may be extended further by adopting well-known techniques for larger context sizes, which NextLevelBERT can then amplify.
For instance, if a basic token-level LM's context size is extended from 512 to 4,096 tokens, this could increase NextLevelBERT's effective context size quadratically up to 16,777,216 tokens.

\section{Experiments}
\begin{table}[htb]
    \centering
    \resizebox{\linewidth}{!}{
        \begin{tabular}{lrrrr}
            \toprule
Dataset  & Pile Books 3 &BookSum &MuLD-Mov &QuALITY\\
\midrule
Training    &156,266&-        &1,256 &2,009   \\
Validation  &19,532 &-        &166   &514   \\
Testing     &19,539       &5,206    &86    &2,086\\

\midrule
\# tokens (avg.) &$\sim$ 125,000 &3,958     &34,706  &5,430  \\
\# tokens (max.) &30,368,036 &42,158    &130,356 &7,992 \\
\bottomrule
        \end{tabular}
    }
    \caption{Overview of the number of documents per dataset. For the BookSum STS task, we test in a zero-shot setting.}
    \label{table:datasets}
\end{table}

To study the overall effectiveness as well as the influence of different chunk encoders and chunk sizes, we pretrain NextLevelBERT on the books3 dataset and then evaluate the model on three downstream tasks: 1) Zero-shot semantic textual similarity (STS) with document vectors, 2) document classification, 3) multiple-choice question answering. We believe that these task types, while not exhaustive, are  representative of common application scenarios for long-document embeddings. Furthermore, they are all story-based, and likely particularly dependent on long-range connections.

\subsection{Pretraining}

\paragraph{Data} The \textbf{books3} dataset is a collection of 197,000 books in plain-text format that is also part of The Pile \citep{gao_pile_2020}, a larger collection of Web text from diverse domains and sources that is frequently used for language modeling. Books are particularly useful due to their length as well as the presence of long-range dependencies that a model may attempt to capture.

\paragraph{Setup}
We pretrain each version of the model for 20 epochs on the encoded books3 dataset for up to five days on four Tesla V100-SXM2-32GB. Pretraining time is shorter for larger chunk sizes. We use a cosine learning rate schedule (max.\ $10^{-4}$) with linear warm-up (first 5\% of the total number of steps) and the AdamW \citep{loshchilov_decoupled_2018} optimizer ($\beta_1=0.9$, $\beta_2=0.999$, weight decay: $0.01$). Unless otherwise stated, we use a global batch size of 512, a masking rate of 15\% and a smooth L1 loss. 
By default, to encode the input chunks into vectors, we use the MiniLM-L6 model from the Sentence Transformers library\footnote{\url{https://www.sbert.net/index.html}}. In practice, we encode the input chunks during preprocessing, since this can be done very efficiently in large batches over multiple GPUs and can be cached for later reuse. 
We initialize the NextLevelModel's weights with the pretrained weights of the chunk encoder, since we found that this is crucial for convergence (see \autoref{pretraining_results}).

\paragraph{Pretrained Chunk Encoder Models} 
To evaluate the choice of chunk encoder, we pretrain several NextLevelBERT versions with different SBERT encoder models \citep{reimers_sentence-bert_2019}. Our focus is on the following three lightweight models with moderate embedding sizes to keep computation costs and memory usage low.

\textbf{(1) MiniLM-L6} \citep{wang_minilm_2020}
is a lightweight BERT-based distilled model with strong performance on the GLUE benchmark. We use the version with six layers.\footnote{\url{https://huggingface.co/nreimers/MiniLM-L6-H384-uncased}}

\textbf{(2) MPNet} \citep{song_mpnet_2020} is an encoder-only model that leverages permuted language modeling and improved position modeling to address drawbacks from previous encoder-only language models. It demonstrates strong performance for sentence embedding tasks. It is almost four times larger than the MiniLM-L6 model in terms of parameters and uses a larger embedding dimensionality of 768. This might be advantageous for modeling longer text, since the additional dimensions can capture more information.

\textbf{(3) DistilRoBERTa} \citep{sanh_distilbert_2019} is a distilled version of RoBERTa-base. This model is smaller than MPNet while also offering a larger embedding dimensionality of 768.

\begin{table*}[h!]
    \centering
    \resizebox{1.0\linewidth}{!}{
        \begin{tabular}{lrrrccccc}
        \toprule
        &  & & &\multicolumn{2}{c}{BookSum}  &\multicolumn{2}{c}{MuLD-Mov} &QuALITY\\
        \cmidrule(lr){5-6}\cmidrule(lr){7-8}\cmidrule(lr){9-9}
        Model \& Chunking &Context &Emb-Dim. &Size (MB) &MRR &HR@10 & Acc &F1 & Acc\\
        \midrule
        NextLevelBERT    \\
        --- Sentence-based &$\sim$12,288 &384 &160 &31.99 &51.11 &80.47 &82.79  &32.12\\
        --- 16 &8,192  &384 &160   &29.78   &49.64  &58.14  &73.53  &24.20\\
        --- 32 &16,384 &384 &160   &24.52   &44.26  &58.14  &73.53  &30.45\\
        --- 64 &32,768 &384 &160   &38.99   &64.89  &59.77  &74.18  &31.58\\
        --- 128 &65,536  &384 &160 &60.27   &81.81  &72.33  &73.67  &32.49\\
        --- 256 &131,072 &384 &160 &\textbf{65.45}   &\textbf{85.57}  &\textbf{83.49}  &\textbf{85.70}  &31.23\\
        --- 512 &262,144 &384 &160 &48.04   &72.13  &64.65  &72.97  &29.03\\
        \midrule
        Avg. SBERT: MiniLM & \\
        --- Sentence-based&N/A &384 &80 &30.94  &50.85  &58.14  &73.53  &32.51\\
        --- 16 &N/A &384 &80 &25.63 &43.64  &58.14  &73.53  &31.93\\
        --- 32 &N/A &384 &80 &38.64 &59.93  &58.14  &73.53  &31.69\\
        --- 64 &N/A &384 &80 &50.35 &73.18  &58.14  &73.53  &31.81\\
        --- 128&N/A &384 &80 &59.37 &80.85  &58.14  &73.53  &31.95\\
        --- 256 &N/A&384 &80 &56.84 &78.49  &58.14  &73.53  &31.89\\
        --- 512&N/A &384 &80 &29.25 &48.98  &58.14  &73.53  &31.67\\
        \midrule
        SBERT: MiniLM &512 &384 &80 &30.87  &48.35  &58.14  &73.53 &31.96\\
        
        Longformer-base &4,096 &768 &595 &\hphantom{0}1.92  &\hphantom{0}4.48   &58.14  &73.53 &26.41\\
        Nomic Embed &8,192 &768 &550 &65.07 &84.90 &60.47 &73.85 &\textbf{36.58}\\
        \bottomrule
        \end{tabular}}
    \caption{Evaluation results (in \%, averaged over five fine-tuning runs). Best results per dataset and metric are presented in bold. The sentence-based NextLevelBERT model's context size is estimated from Books3's average sentence length. For the Chunk Mean approach, there is no overall context limit due to the simple mean-pooling aggregation.}
    \label{table:main_results}
\end{table*}

\subsection{Downstream Evaluation}

We use the following three benchmark datasets.
Additional statistics on the datasets are provided in \autoref{table:datasets}.

\paragraph{BookSum} is an English language narrative summarization dataset that provides human-written summaries at the level of paragraphs, chapters, and entire books \citep{kryscinski_booksum_2021}. We use the chapter-level version from the Huggingface Hub\footnote{\url{https://huggingface.co/datasets/kmfoda/booksum}} and recast the task as a Semantic Textual Similarity one by positing that each chapter should be most similar to its respective summaries rather than to other summaries. The dataset contains 5,206 full-text chapters and 10,910 summaries.

\paragraph{MuLD-Movie Character Classification (MuLD-Mov).}
The MuLD (Multitask Long Document) benchmark is a collection of six datasets for probing long document capabilities \citep{hudson_muld_2022}. Since NextLevelBERT is an encoder-only model, we disregard the tasks requiring generative capabilities and focus on the task of movie character classification. For a given English language movie script and character name, the model needs to determine whether the character is the hero or the villain.

\paragraph{QuALITY} \cite{pang-etal-2022-quality} is a multiple-choice question answering dataset on English language long documents,  including both fiction and non-fiction, with an average length of 5k tokens. For each question, four candidate answers are given. Since the authors have not released labels for the test set, we split the validation set in half at the document level and use one of these halves for testing.

\paragraph{Experimental Setup}

To ensure consistency in the evaluation process, we set it up as similarly as possible across all models. For the BookSum dataset, we generate document embeddings by averaging the output embeddings of pretrained models without any additional fine-tuning. We retrieve the top 10 predictions based on cosine similarity and evaluate them in terms of Mean Reciprocal Rank (MRR) and hit rate. 

For MuLD-Mov.\ and QuALITY, we add a simple classifier head with a hidden layer size of 768 and ReLU activation and subsequently fine-tune the entire model with a learning rate of $5 \times 10^{-5}$. For QuALITY, we concatenate the document vector with an explicit encoding of the question and candidate answer and predict four scores via softmax on the minibatch-level, one per question-answer pair. For MuLD-Mov., we instead concatenate the encoded question: "Is the character <NAME> behaving like a hero or a villain?", where <NAME> is replaced by the actual character's name. We use a sigmoid function (threshold at 0.5) to predict the binary label.

\subsection{Baselines}
We evaluate against the following baselines: \paragraph{Longformer} \citep{beltagy_longformer_2020}, an encoder-only model that uses a local attention window and selected global attention tokens. It is a widely known model for long document tasks. We use the base version with a maximum token length of 4,096.
\paragraph{SBERT} \citep{reimers-gurevych-2019-sentence} is a set of models trained in a cross-encoder setting for STS tasks. We use BERT-based versions, mostly with a 512 token limit to maintain high encoding efficiency. Input text beyond the token limit is truncated.
\paragraph{Avg.\ SBERT} consists of the same SBERT embeddings of the input chunks that NextLevelBERT uses, but simply mean-pools them. Theoretically, the model therefore has access to all information in the document and can directly show the effect of contextualizing the text embeddings with NextLevelBERT. We therefore select the MiniLM model that NextLevelBERT uses in our default setting to encode text chunks.

\paragraph{Nomic Embed} \citep{nussbaum_nomic_2024} is a text embedding model for document lengths up to 8,192 tokens. Based on a BERT architecture, the model is trained via Masked Language Modeling and both unsupervised as well as supervised contrastive textual similarity tasks. We choose this model as a highly competitive baseline with a long context window yet relatively small size at 137M parameters.

\section{Results}

\begin{table*}[h!]
    \centering
        \begin{tabular}{lcccccc}
        \toprule
        &   &\multicolumn{2}{c}{BookSum}  &\multicolumn{2}{c}{MuLD-Mov} &QuALITY\\
        \cmidrule(lr){3-4}\cmidrule(lr){5-6}\cmidrule(lr){7-7}
        Encoder Model &Embed. Dim. &MRR &HR@10 & Acc &F1 & Acc\\
        \midrule
        NextLevelBERT \\
        - MiniLM-256 &384 &\textbf{65.45}   &\textbf{85.57}  &\textbf{83.49}  &\textbf{85.70}  &31.23\\
        - MPNet-256 &768  &\hphantom{0}5.97 &11.03 &58.14 &73.53 &27.74\\
        - DistilRoBERTa-256 &768 &43.05 &66.75 &58.14 &73.20 &29.51\\
        \midrule
        SBERT: MiniLM  &384         &30.87  &48.35  &58.14  &73.53 &31.96\\
        SBERT: MPNet   &768         &44.26 &67.17   &58.14  &73.53  &\textbf{33.05}\\
        SBERT: DistilRoBERTa &768   &27.95 &45.91   &58.14  &73.53 &32.83\\
        \bottomrule
        \end{tabular}
    \caption{Evaluation results (in \%, averaged over five fine-tuning runs) for NextLevelBERT pretrained on the text chunk embeddings of various encoder models. All next-level models use text chunks of size 256.}
    \label{table:encoder_results}
\end{table*}

\subsection{Downstream Task Evaluation}
In \autoref{table:main_results}, NextLevelBERT achieves the best results on both the BookSum semantic similarity and the MuLD-Movie Character task. This shows that language modeling on top of sentence embeddings can indeed enhance document-level embeddings. 

NextLevelBERT and Nomic Embed appear to best handle the BookSum dataset. BookSum full-text chapters are much longer than their respective summaries. Additionally, there are often multiple similar chapters from the same book, which can be difficult to distinguish due to the shared topic and character names. The Avg.\ SBERT approach serves as a strong baseline while the basic SBERT-MiniLM model falls off, indicating that processing the entire text length is crucial for this task. We attribute the poor performance of the Longformer model on BookSum to the fact that it is trained for classification tasks rather than providing general document embeddings.

On the MuLD-Mov.\ dataset, five out of seven NextLevel models as well as Nomic Embed are able to surpass chance level (58\%, i.e., a majority class classifier). All other models fail to do so, even the Avg.\ SBERT baseline, which does well on the other datasets. Notably, our sentence-based and 256er-chunked NextLevelBERT models also outperform the Longformer baseline from \citet{hudson_muld_2022} (82.58 F1), the original benchmark paper, without any task-specific preprocessing. To attain their score, the authors first extracted the most important passages based on character name frequency. We hypothesize that a good performance on MuLD is dependent on a model's ability to process the entire text as well as exchange contextual information internally.

On the QuALITY dataset, Nomic Embed is the strongest model, presumably due to its sufficient context size for this dataset and the fact that it can provide token-level details to answer the questions. The Avg.\ SBERT baselines mostly reach a higher accuracy than NextLevelBERT. Since the chunks that are mean-pooled by Avg.\ SBERT are also the inputs to NextLevelBERT, a lower performance suggests that NextLevelBERT may actively disregard some of the incoming information. This is likely due to the fact that during pretraining, NextLevelBERT learned to focus on reconstructing general events and to ignore local details that would be hard to predict from the chunk-level context. QuALITY consists of questions such as ``Why does Joseph lie about the water supply?'', which require non-trivial consideration of local details. In general, this is a difficult dataset for all the tested models and the results are generally barely above the chance level of 25\%. This is partly reflected in the dataset's leaderboard\footnote{\url{https://nyu-mll.github.io/quality/}}, where end-to-end encoder-only models without additional preprocessing or fine-tuning on related datasets report similar accuracy. The dataset's current SOTA mostly relies on large ($\geq$7B) decoder-only models, training on a large-scale task-specific reading comprehension corpus, and/or retrieval augmentation.

For the same effective context size of 8,192 tokens, Nomic Embed outperforms NextLevelBERT-16 on all datasets. This and the overall results indicate that NextLevelBERT makes heavy use of the pretrained encoder model's ability to contextualize the incoming chunk's text. Smaller chunk sizes, such as 16, likely limit this effectiveness.

\subsection{Impact of Chunking Strategy}
The results in \autoref{table:main_results} illustrate how the optimal choice for dividing the input text into chunks can vary between use cases.
We hypothesize that there is a trade-off between increasing the receptive field by using larger chunks and losing information due to higher levels of compression necessary to encode longer text chunks in fixed-size sentence embeddings. This is supported by the fact that in the cases where the next-level models' context size is not sufficient, performance quickly degrades. A notable exception is NextLevelBERT with sentence-based chunking on the MuLD dataset. With an average context size of $\sim$12k tokens and MuLD's average document length being 34,706 tokens, it still outperforms almost all other models. A possible explanation is that the movie scripts that MuLD consists of are dialogue-heavy and the sentence-based chunking plus the contextualization abilities of the model contribute to successfully predicting the role of the character without seeing the whole document. 

\begin{table}[htb]
    \centering
    \resizebox{\linewidth}{!}{
        \begin{tabular}{lccccc}
        \toprule
        &\multicolumn{2}{c}{BookSum}  &\multicolumn{2}{c}{MuLD-Mov} &QuALITY\\
        \cmidrule(lr){2-3}\cmidrule(lr){4-5}\cmidrule(lr){6-6}
        NextLeveBERT &MRR &HR@10 & Acc &F1 & Acc\\
        \midrule
        with 15\% &\textbf{65.52} &\textbf{85.57} &\textbf{80.93} &\textbf{83.45} &\textbf{31.88}\\
        with 30\% &61.14 &82.10 &69.07 &75.86 &30.08\\
        with 40\% &64.21 &84.81 &72.79 &77.54 &29.53\\
        \midrule
        simpler masking &\hphantom{0}0.77 &\hphantom{0}1.29 &58.14 &73.53 &24.69\\
        \midrule
        random init. &\hphantom{0}1.18 &\hphantom{0}1.78 &58.14 &73.45 &26.44\\
        \bottomrule
        \end{tabular}
}
    \caption{Results for alternative masking rates (1-3), replacing all selected inputs with the [MASK] vector (4), and a randomly initialized model (5). All models use text chunks of size 256. We report the average over five fine-tuning runs.}
    \label{table:design_experiments}
\end{table}

\subsection{Impact of Encoder Model}
Downstream results for NextLevelBERT with different text encoder models are compared in \autoref{table:encoder_results}. We observe that the effectiveness of the original encoder model does not automatically translate to similar outcomes in the next-level setting. Surprisingly, pretraining on MiniLM leads to the best downstream results, although it is the smallest model in terms of parameters as well as embedding dimensionality. A possible explanation is that capturing the intricacies of a 384-dimensional embedding space during pretraining is easier than those of a 768-dimensional space.

\subsection{Impact of Document Length on Model Performance}
\begin{figure*}[t!]
    \begin{subfigure}{\columnwidth}
  \includegraphics[width=\columnwidth]{{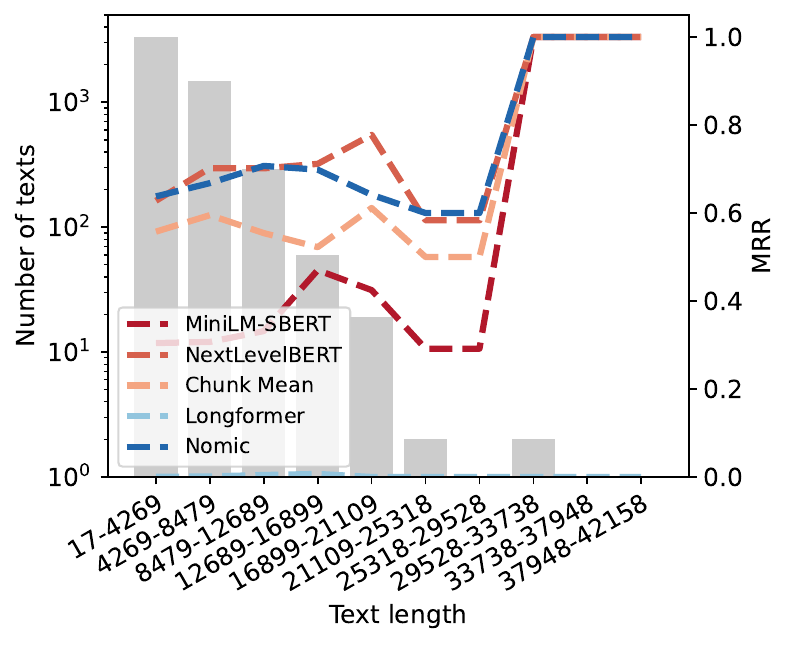}}
  \caption{MRR on all BookSum documents per length}
    \label{fig:mrr_by_length_1}
\end{subfigure}
\begin{subfigure}{\columnwidth}
    \includegraphics[width=\columnwidth]{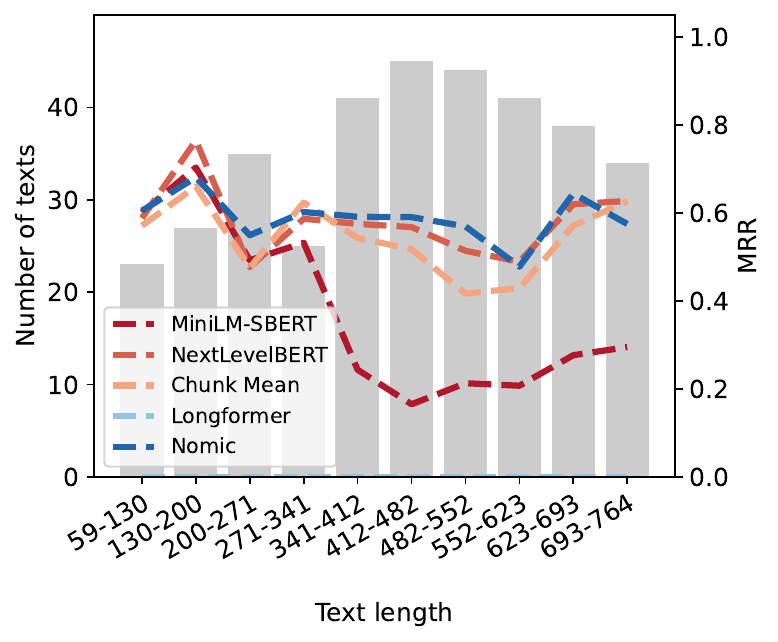}
    \caption{MRR on shorter documents with up to 764 tokens.}
    \label{fig:mrr_by_length_2}
\end{subfigure}

    \caption{MRR on BookSum for different document lengths. NextLevelBERT's performance is particularly strong for longer documents but is able to maintain embedding quality even for short documents. The bars depict the number of samples per length on a log scale.}
    \label{fig:mrr_by_length}
\end{figure*}

To investigate how the models perform with regard to the length of the document, we visualize the MRR on BookSum per document length in \autoref{fig:mrr_by_length}. Here, \autoref{fig:mrr_by_length_1} provides a general plot covering all document lengths, while \autoref{fig:mrr_by_length_2} shows a more fine-grained view of documents shorter than 768 tokens to  study if the design for long documents comes at the cost of degraded performance on shorter texts. NextLevelBERT outperforms the other models, except Nomic Embed, across all document lengths with a margin of up to 20\% MRR. Even on the shorter documents, it generally improves over the other models starting from 341 tokens. On the longest documents, all models exhibit a MRR of 1, but this is due to the low number of samples at these lengths. Notably, Longformer seems unable to generate any useful document embeddings for any sequence length out-of-the-box.

\subsection{Pretraining Effects}
\label{pretraining_results}
\paragraph{Masking Rate}
\citet{wettig-etal-2023-mask} found that BERT's original masking rate of 15\% is not optimal for many pretraining scenarios. They recommend masking more tokens the larger the model is and exclusively replacing tokens with a mask token instead of also retaining the original token or replacing it with a random one at a certain chance. To verify if these findings translate to NextLevelBERT, we compare BERT's original masking strategy versus the recommendations by \citet{wettig-etal-2023-mask} with masking rates between 15-40\%. We train the models for 20 epochs on books3 and show the results in \autoref{table:design_experiments}.
The different masking rates show mixed results, which are dependent on the task.
Interestingly, using only the mask token as replacement for prediction targets had a destructive effect on our model's downstream performance, preventing it from learning useful representations. 

\paragraph{Token-Pretrained vs Random Initialization}

To measure the effect of initializing NextLevelBERT with the chunk encoder weights,  we train two versions of the same model, one initialized with the weights of the sentence encoder and one with random initialization. We compare their downstream performance in \autoref{table:design_experiments}.
For the MuLD-Movie and the QuALITY tasks, the model with random initialization does not score above chance level. On BookSum it also underperforms, indicating that it was not able to learn meaningful representations.
Together, these results reveal that initializing NextLevelBERT with the chunk encoder weights is essential to its downstream capabilities.

\section{Related Work}
\label{sec:related_work}
\subsection{Efficient Attention}
Numerous approaches have been proposed to increase the context length by making the attention mechanism more scalable, typically by sparsifying or approximating the attention operation: \citet{beltagy_longformer_2020} use a combination of local and global attention to circumvent the quadratic scaling. They train the Longformer, a BERT-based LM that can incorporate a context size of up to 4,096 tokens, showing improved results on long document classification. Building on the Longformer, \citet{xiao_primera_2022} pack multiple documents into the same sequence and mask salient sentences to generate multi-document summaries. The Poolingformer \citep{zhang_poolingformer_2021} adds a second level of pooling-combined attention to achieve linear computational complexity, while the Reformer \citep{kitaev_reformer_2019} proposes attention with locality-sensitive hashing to reduce the attention operation to linearithmic time. Finally, HyperAttention \citep{han_hyperattention_2023} efficiently approximates the attention matrix and enables computing attention in near-linear time.

\subsection{Hierarchical Approaches for Long Contexts}
Hierarchical approaches usually divide the input text into text chunks. These are then processed in parallel and aggregated at different levels. Hierarchical models thus enable large context sizes through high parallelization and compressed sequence lengths.
After the advent of Transformer-based language models, hierarchical Transformer architectures have been proposed for (long) document classification \citep{pappagari_hierarchical_2019, dai-etal-2022-revisiting}. They usually leverage a pretrained Transformer to encode smaller chunks of text into vector representations. These are then jointly provided to another Transformer model, which is trained to predict the document class.

There are several extractive summarization approaches that adopt related kinds of hierarchical architectures \citep{zhang_hibert_2019, xu_unsupervised_2020, ruan_histruct_2022}. They are all based on a two-level BERT model \citep{devlin-etal-2019-bert} that first encodes sequences of tokens into sentence vectors and then adds another module that operates on the sentence level. \citet{zhang_hibert_2019} pretrain their model HIBERT on a large document corpus and mask entire sentences. Unlike NextLevelBERT, HIBERT inserts the mask on a token level, thus leaking information on the masked sentences' length. It also includes a decoder component that takes the masked sentences' representations and reconstructs the original sentence tokens. This results in semantically equivalent sentences that differ on the token level being disregarded and the task thus remains next-token prediction, albeit with the local sentence context removed. They also restrict each input to 30 sentences, limiting long-range dependencies. Building on HIBERT, \citet{xu_unsupervised_2020} pretrain a hierarchical BERT model, but mask at the encoded sentence level instead of at the token level. They pretrain on their summarization data instead of a large corpus of general documents and also decode the sentence vectors into a sequence of tokens with a next-token prediction objective. Lastly, \citet{ruan_histruct_2022} do not pretrain their model at all. They focus on adding structural information to HIBERT by adapting the positional embeddings based on the input text's structure. Thus, all variations of these hierarchical BERT models are designed solely for extractive summarization, predict the masked sentences at a token level, and have not been evaluated for other tasks, domains or with regard to long contexts. They hence differ from NextLevelBERT in both their design and application.

\subsection{Embedding-based Loss Objectives}
The process of encoding sequences of data and predicting the next item in the sequence's latent representation has also been used in other domains. To generate images autoregressively, \citet{lee_autoregressive_2022} use a variation of a Vector-Quantized Variational Autoencoder (VAE) to encode sequences of image patches into discretized codebook representations. Their model learns to predict the latent representation of the next patch in the sequence. The decoder of the VAE then outputs pixel-level images from the generated latent patch sequence. In reinforcement learning, \citet{lee_autoregressive_2022} use cubic masking patches to corrupt video sequences as an auxiliary task for learning a state representation model. The model learns to reconstruct the uncorrupted videos' latent representations. Similar approaches have also been suggested in \citet{hafner_dream_2020, hafner_mastering_2022}. In NLP, sentence embedding approaches, such as DeCLUTR \citep{giorgi_declutr_2021} and Denoising Autoencoder with Transformers \citep{moradshahi-etal-2023-zero}, learn to generate text embeddings for semantic similarity via a self-supervised loss on the embedding space. DeCLUTR samples text pairs that are adjacent, overlapping, or where one sample subsumes the other to obtain positive pairs for a contrastive loss. The Denoising Autoencoder corrupts input sentences and learns to reconstruct the original sentence.

\section{Conclusion}
Successfully modeling long-range dependencies is an important step towards building language models with general language understanding. Inspired by the idea of modeling language at a coarser resolution than that of tokens, we have proposed NextLevelBERT, and transferred masked language modeling from the token level to general text representation vectors. We pretrained NextLevelBERT on a large collection of books and evaluated the resulting text representations on three diverse downstream tasks in a long document setting. Our results show that next-level masked language modeling leads to sensibly contextualized text chunk representations and generates useful long-document vectors, especially for tasks that do not require specific token-level information. Additionally, NextLevelBERT is very parameter-efficient, using only a third of the parameters of the Nomic Embed model while maintaining similar performance.

In the future, combining the predictions of a next-level model with a token-level component based on the required granularity could be a way of preserving token-level details. Furthermore, pretraining NextLevelBERT on data from more diverse domains could improve its performance in general and extend its usefulness to a wider range of tasks. Lastly, while not straightforward, adapting decoder-only models to the next level could translate similar gains as in the MLM setup to generative language models.

\section*{Limitations}
Our NextLevelBERT model currently has limited usefulness for tasks from domains and styles dissimilar to those covered by the books3 dataset, since it is not pretrained on other data. Our experiments thus far only cover the English language, and further languages remain to be explored.
Encoding the input text chunks can either be conducted during preprocessing or on-the-fly during training. The former offers increased parallelization and time efficiency, but the entire pretraining dataset needs to be saved as vectors. This can take up a lot of space, especially for large embedding dimensionalities. The latter does not require additional storage but significantly slows down training, since batch sizes in the encoder do not match those of the NextLevel component. This means that NextLevelBERT is limited to encoder models with relatively few parameters and a small embedding dimensionality.

\section*{Ethical Considerations}
The Pile and the books3 dataset have become the subject of controversy due to the inclusion of copyrighted materials, which are widely used to train large language models. While there is no final verdict on whether training Generative AI models on such data constitutes fair use, there is a growing consensus that the wishes of authors to have their work excluded from Generative AI model training ought to be respected. Our study uses this data exclusively for academic purposes, to assess the merits of masked language modeling at a higher level on longer documents. Our model is not generative and it is hence impossible to use it to reproduce non-trivial verbatim snippets of text from the books3 dataset. 
Moreover, we will take care not to share any such data as part of our code repository.

As a text encoder model pretrained without specific anti-bias measures and building on other text encoder representations, NextLevelBERT is likely to exhibit various kinds of social biases.
\section*{Acknowledgements}
Gerard de Melo received funding from The Goldman Sachs Group, Inc., New York, NY, USA. Christoph Hönes's research is funded by SAP SE, Germany.

\bibliography{anthology,references,custom}

\end{document}